# Semantic Adversarial Examples

Hossein Hosseini      Radha Poovendran

Network Security Lab (NSL)

Department of Electrical Engineering, University of Washington, Seattle, WA

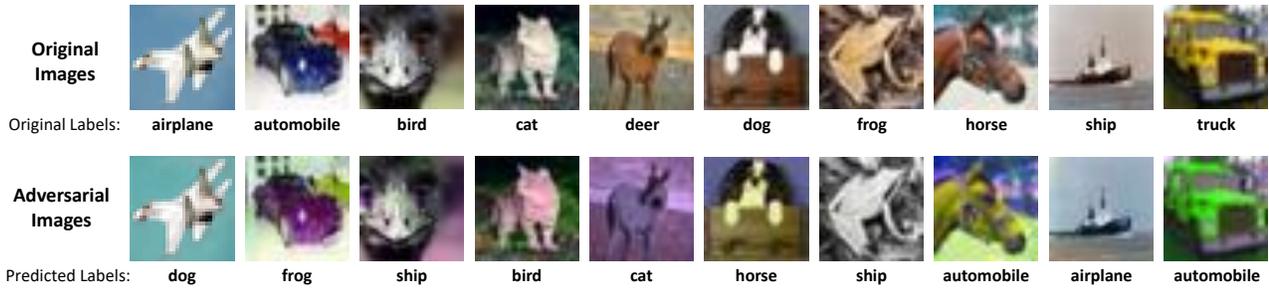

Figure 1: Samples of CIFAR10 original images (top) and semantic adversarial examples (bottom) on VGG16 network. Adversarial images are generated by converting original images into the HSV color space and randomly shifting the Hue and Saturation components, while keeping Value the same. All images in first row are correctly classified by the model.

## Abstract

*Deep neural networks are known to be vulnerable to adversarial examples, i.e., images that are maliciously perturbed to fool the model. Generating adversarial examples has been mostly limited to finding small perturbations that maximize the model prediction error. Such images, however, contain artificial perturbations that make them somewhat distinguishable from natural images. This property is used by several defense methods to counter adversarial examples by applying denoising filters or training the model to be robust to small perturbations.*

*In this paper, we introduce a new class of adversarial examples, namely "Semantic Adversarial Examples," as images that are arbitrarily perturbed to fool the model, but in such a way that the modified image semantically represents the same object as the original image. We formulate the problem of generating such images as a constrained optimization problem and develop an adversarial transformation based on the shape bias property of human cognitive system. In our method, we generate adversarial images by first converting the RGB image into the HSV (Hue, Saturation and Value) color space and then randomly shifting the Hue and Saturation components, while keeping the Value component the same. Our experimental results on CIFAR10 dataset show that the accuracy of VGG16 network on adversarial color-shifted images is $5.7\%$.*

## 1. Introduction

Image classifiers are vulnerable to adversarial inputs, i.e., it is possible to carefully modify an image such that the model will classify it into a wrong label, while a human observer perceives the original object [1, 2]. The existence of adversarial examples is intriguing from learning perspective, since models that outperform human in classifying natural images can be so easily fooled by adding a hardly visible perturbation. Also from the security perspective, such attacks pose a major threat, as machine learning systems are being increasingly integrated into critical and security-sensitive applications, such as autonomous cars, medical diagnosis, and banking.

Generating adversarial examples has been mostly limited to finding small perturbations that maximize the model prediction error [3–6]. Such modified images, however, contain artificial perturbations that make them somewhat distinguishable from natural images. This property is used by several defense methods to make deep learning models robust against small perturbations [7, 8] or to map the perturbed image back into the space of natural images by applying preprocessing filters [9, 10].

In practice, however, the adversary may not be constrained with slightly modifying the image. That is, the adversary may perturb the image a lot, but in such a way that the modified image semantically represents the same object as the original image (because otherwise, we cannot expect the model to classify it correctly). To construct such images, we need to identify the types of transformations that human vision is invariant to and investigate how do start-of-the-art deep learning models compare to humans.

One property of human cognitive system is the "shape bias", i.e., when assigning a name to new items, humans weight shape more heavily than they do other dimensions of perceptual similarity, such as size or texture [11]. Convolutional Neural Networks (CNNs) are similarly designed to take into account the spatial structure of image data. In fact, models trained on ImageNet are shown to display shape bias as well [12].

In this paper, we make the following contributions.

- We introduce a new class of adversarial examples, namely *semantic adversarial examples*, as images that are arbitrarily perturbed to fool the model, but semantically represent the original objects. We formulate the problem as a constrained optimization problem, requiring the modified image to be smooth and natural-looking, so as to be undetectable by current defense methods.

- We propose a method for generating semantic adversarial images, based on the shape bias property of human cognitive system. In our method, we first convert the image from RGB into the HSV color space, composed of Hue, Saturation and Value color channels. We then randomly shift the hue and saturation components, while keeping the value the same. This approach generates images that contain the original object with different colors and colorfulness.

- We perform the experiments on CIFAR10 dataset and VGG16 network. The results show that the model accuracy on adversarial color-shifted images is $5.7\%$. Figure 1 shows samples of CIFAR10 original images and their corresponding modified images that fool the VGG16 model. As can be seen, modified images represent the same object as original images. We also apply the attack on the state-of-the-art robust model against adversarial examples [8] and show that the model accuracy drops to $8.6\%$ when tested on adversarial color-shifted images. The code for generating adversarial color-shifted images is available at `https://github.com/HosseinHosseini/Semantic-Adversarial-Examples`.

The rest of the paper is organized as follows. Section 2 presents the problem of generating semantic adversarial examples. In section3, we first provide a background on shape bias property and HSV color space and then propose a method for generating adversarial color-shifted images. Section 4 presents the experimental results. Section 5 reviews the related works and Section 6 concludes the paper.

## 2. Problem Statement

### 2.1. Adversarial Examples

In this paper, we consider the misclassification attack. Current techniques for generating adversarial examples try to find a perturbation that maximizes the network prediction error. Let $F$ be the machine learning classifier and $x$ be the given image. The adversarial perturbation is typically found by solving the following optimization problem [2]:

$$\min \|\delta\| \qquad (1)$$
$$\text{s.t. } F(x + \delta) \neq F(x).$$

The added perturbation in adversarial examples is usually small and, hence, the modified image is likely to belong to the same class as the original image. The image, however, does contain an artificial perturbation that makes it somewhat distinguishable from natural images. This property is used by several defense methods to counter adversarial examples by explicitly applying denoising operations [9, 10] or training the model to do so implicitly [7, 8].

### 2.2. Semantic Adversarial Examples

In practice, the adversary may not be constrained with slightly modifying the image. That is, the image can be modified by any transformation, conditioned that the transformation preservers the semantics of the image. Let $\Omega$ be the human vision system. The problem of generating semantic adversarial examples is stated as follows:

$$\text{find } x^* \qquad (2)$$
$$\text{s.t. } \Omega(x^*) = \Omega(x) \text{ and } F(x^*) \neq F(x).$$

The problem (2) can be seen as mapping any given image into the space of natural images that are misclassified by the model, but contain the original object. In this sense, wrongly-classified clean images are adversarial examples with zero perturbation.

Identifying and studying adversarial transformations is important from the learning perspective, since it helps to investigate how the model compares to human visual system and also to analyze the model generalization performance. Moreover, such adversarial transformations will be able to evade state-of-the-art defense methods that try to reverse the added perturbation. Therefore, it is important also from the security perspective to identify the attack space and develop more robust defense mechanisms.

## 3. Proposed Method

In this section, we first review the shape bias property of human cognitive system. We then provide a background on HSV color space and finally propose a method for generating semantic adversarial examples by shifting the image color components.

## 3.1. Shape Bias Property of Human Cognitive System

To construct semantic adversarial examples, we need to identify the properties of human vision system. One such property is the "shape bias," stating that humans prefer to categorize objects according to their shape rather than color [11]. Therefore, we expect machine learning models to also correctly classify images that contain the original object with different colors. In the following, we propose a method to generate such images.

## 3.2. HSV Color Space

HSV (Hue, Saturation and Value) is an alternative to RGB (red, green and blue) color space and is known to more closely represent the way human vision perceives color [13]. The hue channel corresponds to the color's position on the color wheel. As hue increases from 0 to 1, the color transitions from red to orange, yellow, green, cyan, blue, magenta, and finally back to red. Saturation measures the colorfulness, i.e., setting saturation to 0 yields a gray-scale image and increasing it to 1 generates the most colorful image with same colors. Value shows the brightness, which is maximum value of red, green and blue components.

## 3.3. Color-Shifted Images as Semantic Adversarial Examples

HSV can be seen as a color space, in which color components (hue and saturation) are decoupled from the object structure (brightness). Therefore, by changing the hue and saturation components and keeping the value the same, we can generate images that contain the original object with different color and colorfulness. Let $x_H, x_S$ and $x_V$ respectively denote the hue, saturation and value components of image $x$. The problem of generating semantic adversarial examples by changing the image color can be stated as follows:

$$\text{find } x^* \qquad (3)$$
$$\text{s.t. } x_V^* = x_V \text{ and } F(x^*) \neq F(x).$$

For solving (3), we can generate random images, set their value component to $x_V$ and then choose the ones that are misclassified by the model. Equally, we can start from the given image and randomly perturb the hue and saturation components in such a way that the modified image can fool the model. These methods, however, will generate images with visible noise.

In order to generate smooth and natural-looking images, we *shift* the hue and saturation components of all pixels by the same amount. This approach generates an image where all pixels are equally colored if the shift in saturation is positive or decolored if the shift is negative. The color of all pix-els (the hue component) is also shifted by a fixed amount. We call such images as *color-shifted images*.

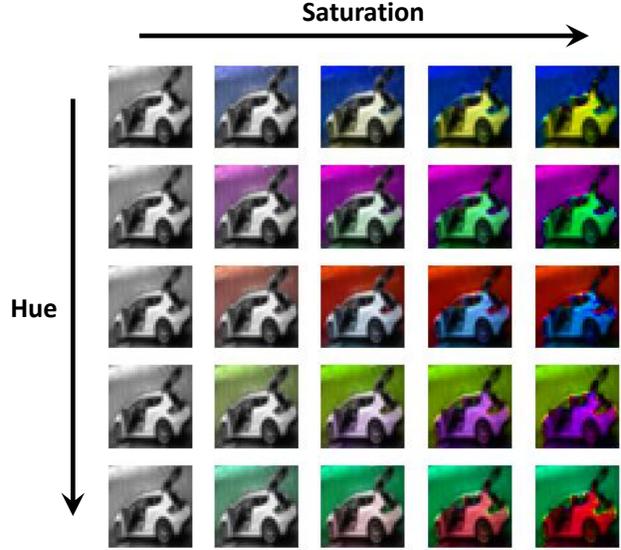

Figure 2: Illustration of shifting color components (hue and saturation) in HSV color space on a sample image of CIFAR10 dataset. The center image is the original one. Shifting hue and saturation components changes the color and colorfulness, respectively. As can be seen, the original object is recognizable in all images.

We also note that significantly decreasing or increasing the saturation component generates gray-scale or too colorful images, respectively, and hence causes the image to be less like a natural color image. To generate better looking images, we add a requirement that the saturation component be minimally shifted. Figure 2 shows a sample image of CIFAR10 dataset and several color-shifted images. The center image is the original one. As can be seen, images differ in color and colorfulness across variations in hue and saturation components, respectively. Also, while the original object is recognizable in all images, images closer to the middle column look more like natural color images.

Let $\delta_H$ and $\delta_S$ denote the shifts in hue and saturation components, respectively. Adversarial color-shifted images can be generated by solving the following problem:

$$\min |\delta_S| \qquad (4)$$
$$\text{s.t. } \begin{cases} x_H^* = (x_H + \delta_H) \bmod 1 \\ x_S^* = \text{clip}(x_S + \delta_S, 0, 1) \\ x_V^* = x_V \end{cases}$$
$$\text{and } F(x^*) \neq F(x).$$

Note that $\delta_H$ and $\delta_S$ are scalars. The color in hue component changes in a circle, i.e., hue of 1 is equal to hue of 0. Hence, we compute the modulo of hue component with 1 to map it to $[0, 1]$. The saturation component, however, should be clipped to the interval of $[0, 1]$.

### 3.4. Algorithm

To solve (4), we do a random search over parameters $\delta_H$ and $\delta_S$, and obtain the modified image as described in first constraint of (4). We continue generating color-shifted images until the modified image is misclassified by the model or the maximum number of trials, denoted by $N$, is reached. The shift in hue component is chosen as $\delta_H \sim U(0, 1)$, where $U(a, b)$ denotes uniform distribution in $[a, b]$. However, to find adversarial images with smaller saturation shift, $\delta_S$ is chosen as $\delta_S \sim U(-\frac{i}{N}, \frac{i}{N})$, where $i$ is the iteration number. That is, we start with $\delta_S = 0$ and linearly increase the interval after each trial. Algorithm 1 describes the method.

---

**Algorithm 1** Generating Adversarial Color-shifted Images

1: **Input:** Classifier $F$, Image $x$, Maximum number of trials $N$
2: **Output:** Adversarial color-shifted image $x^*$ or $\emptyset$
3: $x_H, x_S, x_V \leftarrow$ Hue, Saturation and Value components of image $x$, respectively.
4: $x^* \leftarrow x$
5: **for** $i = 0, ..., N-1$ **do**
6: $\quad \delta_H \leftarrow$ a number uniformly chosen in $[0, 1]$
7: $\quad \delta_S \leftarrow$ a number uniformly chosen in $[-\frac{i}{N}, \frac{i}{N}]$
8: $\quad x_H^* = (x_H + \delta_H) \bmod 1$
9: $\quad x_S^* = \text{clip}(x_S + \delta_S, 0, 1)$
10: $\quad$ **if** $F(x^*) \neq F(x)$ **then**
11: $\quad\quad$ **return** $x^*$
12: $\quad$ **end if**
13: **end for**
14: **return** $\emptyset$

---

## 4. Experimental Results

Experiments are performed on image dataset CIFAR10, which consists of natural color images in 10 classes of airplane, automobile, bird, cat, deer, dog, frog, horse, ship and truck [14]. We apply the attack on the pretrained VGG16 network [15], a robust network proposed by Madry et al. [8], and the VGG16 network trained with both original and color-shifted images. In experiments, we set maximum number of trials $N = 1000$.

Table 1 provides the results for different models. The accuracy of pretrained VGG16 network drops to $5.7\%$ when tested on adversarial color-shifted images. Figure 1 shows samples of original images and their corresponding color-shifted images that are misclassified by the model. As can be seen, the original object is recognizable in all of the adversarial images. Figure 3 shows attack success rate versus number of trials. For more than $14\%$ of images, the model can be fooled by trying only one modified image, obtained by randomly shifting only the hue component. The results show that although the model achieves accuracy of $93.6\%$ on test data, it is fragile to images with maliciously-shifted color components.

Table 1: Accuracy of different CNNs on test data and adversarial color-shifted images. The VGG-augmented is a VGG16 network trained with both original and color-shifted images.

| Network | Accuracy on test images | Accuracy on adversarial color-shifted images |
|---|---|---|
| Pretrained VGG16 [15] | 93.6% | 5.7% |
| Madry et al. Model [8] | 87.3% | 8.4% |
| VGG-augmented | 89.9% | 69.1% |

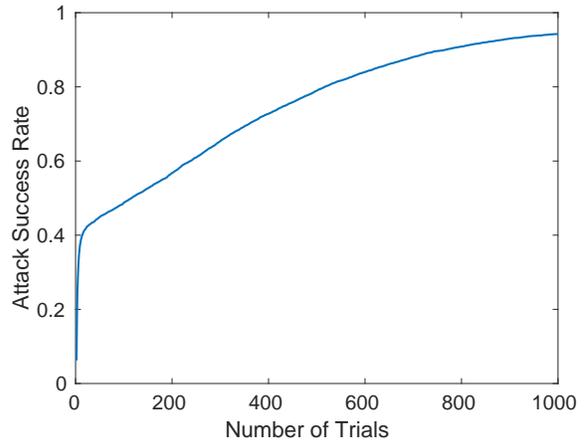

Figure 3: Attack success rate versus number of trials. For more than $14\%$ of images, the model is fooled after trying only one color-shifted image, obtained by randomly shifting only the hue component.

We also applied targeted attack on VGG16 network. In our method, the attack space is limited, since we search over only two parameters. Nevertheless, we could achieve $35.4\%$ success rate on images that are correctly classified by model, i.e., adversarial color-shifting can change the model prediction to an average of more than three classes. Figure 4 shows sample images of CIFAR10 dataset, each with three color-shifted versions classified into different labels.

The model proposed by Madry el al. yields the state-of-the-art results against adversarial examples [16], by providing robustness against worse-case perturbations [8]. We apply our attack also on this model to ensure that our method does not add noise-like perturbation to the image. We observed that the accuracy of the robust CNN is $8.4\%$ on adversarial color-shifted images, implying that even if the model is robust to perturbations around data points, it does not provide robustness to semantic adversarial examples.

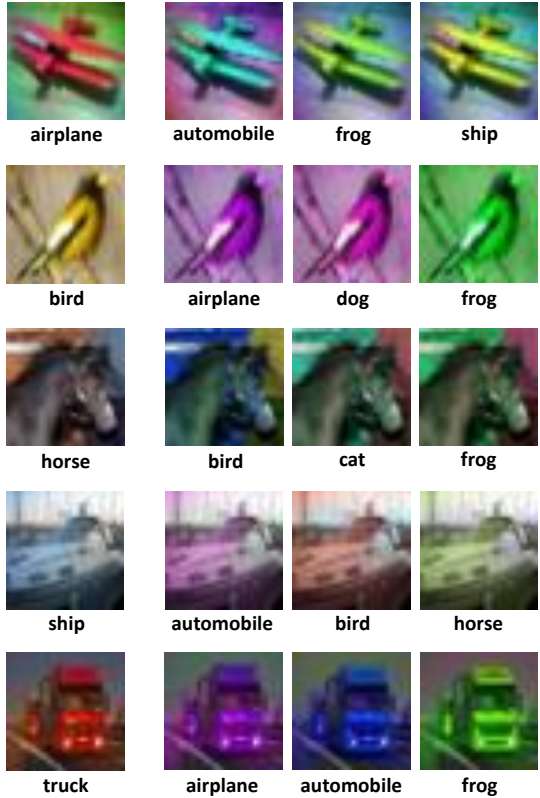

Figure 4: Sample images of CIFAR10 dataset, along with three color-shifted versions classified into different labels by VGG16 network. The leftmost-column shows original images.

We also applied the attack on a VGG16 network which, at each epoch, is trained with original and color-shifted images with $\delta_H \sim U(0, 1)$ and $\delta_S \sim U(-1, 1)$. The accuracy on adversarial color-shifted images is $69.1\%$, indicating that the model shows more robustness when trained with same types of images. However, as pointed out in [17], robustness achieved by data augmentation may not be an indication that the model has learned higher level semantic features in the dataset. That is, the network is likely to be vulnerable to other types of semantic adversarial examples. We will explore other attack and defense methods in future works.

## 5. Related Work

Over the past several years, many techniques have been proposed to generate adversarial examples by finding a small perturbation that can fool the model. Such techniques, including Fast Gradient Sign Method (FGSM) [3], Deep-Fool [4], Projected Gradient Descent (PGD) [5], and Carlini and Wagner attacks [6], usually involve an optimization problem that solves for a perturbation that maximizes the model prediction error. In this paper, we proposed an image transformation that may introduce large pixel-wise perturbation, but preserves the semantics of the image. Our proposed transformation has tunable parameters that can be searched over to generate adversarial images.

It has been shown that CNNs trained with clean images generalize poorly to images with reversed brightness, called negative images [18]. Compared to color-shifted images, negative images are less recognizable to humans [19]. Moreover, image complementing is a fixed transformation and cannot be adjusted to force the model to misclassify images. In [17], Fourier filtering methods are used to slightly modify images such that they contain the same high level abstractions but exhibit different surface statistical regularities. They showed that, in the worst case, the CNN accuracy reduces more than $20\%$ when tested on such images. In comparison, our method introduces large perturbation to the image and reduces model accuracy by about $90\%$.

In [20], the authors showed that adversarial translation and rotation can fool CNNs. The rotated images, however, may look unnatural to humans, due to the black borders and the existence of tilted objects. In [21], the authors considered the problem of generating natural adversarial examples using generative adversarial networks. Their proposed method generates images that look natural, but may belong to a different class than the original class. Hence, such images do not fit into our definition of semantic adversarial examples. In [22], the authors suggested that instead of small or imperceptible perturbations, the adversary may opt for more effective but noticeable perturbations, and proposed a method to create adversarial patches. Our method also generates images with large modification, however in such a way that the perturbation is not visible, i.e., the modified image is smooth and natural-looking.

## 6. Conclusion

In this paper, we introduced Semantic Adversarial Examples as images that semantically represent the original object, but are misclassified by the model. We proposed a method for generating such images by shifting the color components of the image in HSV color space, and showed that the generated images are smooth and natural-looking. Our experimental results on CIFAR10 dataset show that the accuracy of state-of-the-art CNNs on adversarial color-shifted images is less than $10\%$.

## Acknowledgments

This work was supported by ONR grants N00014-14-1-0029 and N00014-16-1-2710, ARO grant W911NF-16-1-0485 and NSF grant CNS-1446866.

# References

[1] B. Biggio, I. Corona, D. Maiorca, B. Nelson, N. Šrndić, P. Laskov, G. Giacinto, and F. Roli, "Evasion attacks against machine learning at test time," in *Joint European conference on machine learning and knowledge discovery in databases*, pp. 387–402, Springer, 2013.

[2] C. Szegedy, W. Zaremba, I. Sutskever, J. Bruna, D. Erhan, I. Goodfellow, and R. Fergus, "Intriguing properties of neural networks," *arXiv preprint arXiv:1312.6199*, 2013.

[3] I. J. Goodfellow, J. Shlens, and C. Szegedy, "Explaining and harnessing adversarial examples," *arXiv preprint arXiv:1412.6572*, 2014.

[4] S. M. Moosavi Dezfooli, A. Fawzi, and P. Frossard, "Deepfool: a simple and accurate method to fool deep neural networks," in *Proceedings of 2016 IEEE Conference on Computer Vision and Pattern Recognition (CVPR)*, no. EPFL-CONF-218057, 2016.

[5] A. Kurakin, I. Goodfellow, and S. Bengio, "Adversarial machine learning at scale," *arXiv preprint arXiv:1611.01236*, 2016.

[6] N. Carlini and D. Wagner, "Towards evaluating the robustness of neural networks," in *Security and Privacy (SP), 2017 IEEE Symposium on*, pp. 39–57, IEEE, 2017.

[7] F. Tramèr, A. Kurakin, N. Papernot, D. Boneh, and P. McDaniel, "Ensemble adversarial training: Attacks and defenses," *arXiv preprint arXiv:1705.07204*, 2017.

[8] A. Madry, A. Makelov, L. Schmidt, D. Tsipras, and A. Vladu, "Towards deep learning models resistant to adversarial attacks," *arXiv preprint arXiv:1706.06083*, 2017.

[9] Y. Song, T. Kim, S. Nowozin, S. Ermon, and N. Kushman, "Pixeldefend: Leveraging generative models to understand and defend against adversarial examples," *arXiv preprint arXiv:1710.10766*, 2017.

[10] C. Guo, M. Rana, M. Cissé, and L. van der Maaten, "Countering adversarial images using input transformations," *arXiv preprint arXiv:1711.00117*, 2017.

[11] B. Landau, L. B. Smith, and S. S. Jones, "The importance of shape in early lexical learning," *Cognitive development*, vol. 3, no. 3, pp. 299–321, 1988.

[12] S. Ritter, D. G. Barrett, A. Santoro, and M. M. Botvinick, "Cognitive psychology for deep neural networks: A shape bias case study," *arXiv preprint arXiv:1706.08606*, 2017.

[13] A. Koschan and M. Abidi, *Digital color image processing*. John Wiley & Sons, 2008.

[14] A. Krizhevsky, "Learning multiple layers of features from tiny images," 2009.

[15] K. Simonyan and A. Zisserman, "Very deep convolutional networks for large-scale image recognition," *arXiv preprint arXiv:1409.1556*, 2014.

[16] A. Athalye, N. Carlini, and D. Wagner, "Obfuscated gradients give a false sense of security: Circumventing defenses to adversarial examples," *arXiv preprint arXiv:1802.00420*, 2018.

[17] J. Jo and Y. Bengio, "Measuring the tendency of cnns to learn surface statistical regularities," *arXiv preprint arXiv:1711.11561*, 2017.

[18] H. Hosseini and R. Poovendran, "Deep neural networks do not recognize negative images," *arXiv preprint arXiv:1703.06857*, 2017.

[19] R. J. Phillips, "Why are faces hard to recognize in photographic negative?," *Perception & Psychophysics*, vol. 12, no. 5, pp. 425–426, 1972.

[20] L. Engstrom, D. Tsipras, L. Schmidt, and A. Madry, "A rotation and a translation suffice: Fooling cnns with simple transformations," *arXiv preprint arXiv:1712.02779*, 2017.

[21] Z. Zhao, D. Dua, and S. Singh, "Generating natural adversarial examples," *arXiv preprint arXiv:1710.11342*, 2017.

[22] T. B. Brown, D. Mané, A. Roy, M. Abadi, and J. Gilmer, "Adversarial patch," *arXiv preprint arXiv:1712.09665*, 2017.

[23] A. Krizhevsky, I. Sutskever, and G. E. Hinton, "Imagenet classification with deep convolutional neural networks," in *Advances in neural information processing systems*, pp. 1097–1105, 2012.

[24] M. Paulin, J. Revaud, Z. Harchaoui, F. Perronnin, and C. Schmid, "Transformation pursuit for image classification," in *Proceedings of the IEEE Conference on Computer Vision and Pattern Recognition*, pp. 3646–3653, 2014.

[25] K. Chatfield, K. Simonyan, A. Vedaldi, and A. Zisserman, "Return of the devil in the details: Delving deep into convolutional nets," *arXiv preprint arXiv:1405.3531*, 2014.

[26] S. Hauberg, O. Freifeld, A. B. L. Larsen, J. Fisher, and L. Hansen, "Dreaming more data: Class-dependent distributions over diffeomorphisms for learned data augmentation," in *Artificial Intelligence and Statistics*, pp. 342–350, 2016.